\documentclass[10pt,twocolumn,letterpaper]{article}

\usepackage[pagenumbers]{wacv} 

\definecolor{wacvblue}{rgb}{0.21,0.49,0.74}
\usepackage[pagebackref,breaklinks,colorlinks,allcolors=wacvblue]{hyperref}
\usepackage[accsupp]{axessibility}

\def\wacvPaperID{2821} 
\def\confName{WACV}
\def\confYear{2027}

\title{Biological Congruence: Evaluating Microscopy Generation Across Representations}

\author{
Duway Nicolas Lesmes-Leon\textsuperscript{1,2}\thanks{These authors contributed equally to this work.} \quad
Shumpei Takezaki\textsuperscript{3}\footnotemark[1] \quad
Fabian Schmeisser\textsuperscript{1,2} \quad
Ryoma Bise\textsuperscript{3}
\\[0.5em]
Gillian Lovell\textsuperscript{4} \quad
Bianca Migliori\textsuperscript{4} \quad
Andreas Dengel\textsuperscript{1,2} \quad
Sheraz Ahmed\textsuperscript{1}
\\[1em]
\textsuperscript{1}German Research Center for Artificial Intelligence (DFKI), Kaiserslautern, Germany\\
\textsuperscript{2}RPTU Kaiserslautern-Landau, Kaiserslautern, Germany\\
\textsuperscript{3}Kyushu University, Fukuoka, Japan\\
\textsuperscript{4}Sartorius Corporate, Royston, UK
\\[0.5em]
{\small\ttfamily
lleon@rptu.de \quad
shumpei.takezaki@human.ait.kyushu-u.ac.jp
}
}

\begin{document}
\maketitle
\begin{abstract}
Generative models increasingly produce realistic microscopy images, yet conventional evaluation primarily characterizes visual fidelity and distributional coverage rather than the biological information represented by synthetic data. We introduce \textbf{biological congruence} as an evaluation principle describing whether biologically relevant information forms a coherent description across representations and remains consistent with its empirical biological context. To study this principle, we develop a text-conditioned diffusion framework that jointly generates microscopy images and instance masks, exposing the same generated biological state through conditioning, visual appearance, and instance structure. Across prompt--mask, prompt--image, image--mask, and population-level analyses, we find strongly feature-dependent behavior: density and confluency are consistently preserved, morphology exhibits recurrent disagreement, while nearest-neighbor distance reveals that strong image--mask agreement can coexist with weaker conditioning and population fidelity. Stress testing increasingly unsupported phenotype combinations further shows that target controllability can increase while untargeted phenotypes reorganize and the complete generated state moves toward configurations better supported by real data. These findings demonstrate that generative quality, controllability, cross-representation agreement, and population fidelity capture complementary properties, motivating biological congruence as a relational perspective for evaluating scientific generative models.
\end{abstract}
    
\section{Introduction}
\label{sec:intro}

Generative models are increasingly capable of producing realistic microscopy images and offer new opportunities for data augmentation, simulation, and modeling of cellular phenotypes \cite{GAN_review_plos}. In biological applications, however, visual realism is only one requirement. A generated sample may also be expected to represent a particular cell type, acquisition state, morphology, spatial organization, or other experimentally relevant phenotype. Evaluating biological generation therefore requires asking not only whether synthetic images resemble the real distribution, but whether the biological information they are intended to represent is preserved \cite{trust_gap}.

Conventional generative metrics such as Fréchet Inception Distance (FID), precision, and recall provide useful measures of distributional similarity, fidelity, and coverage \cite{FID,recall_precision}. Biological generative studies increasingly complement them with domain-specific measurements, including phenotype distributions, morphological descriptors, perturbation recovery, and downstream predictive tasks \cite{bio_eval,custom_FID}. However, these measurements do not directly ask whether different representations of the same generated sample support a coherent biological state. This leaves a distinct question: \emph{do different representations of the same generated sample agree on the biological state they describe, and is that state consistent with its biological context?}

We refer to this property as \textit{biological congruence}: the extent to which biologically relevant information forms a coherent description across generated representations and remains consistent with the biological context it is intended to represent. Rather than proposing an aggregate metric, we treat biological congruence as an evaluation principle based on complementary relationships among representations and their agreement with empirical biological context. Agreement in one relationship does not necessarily imply agreement in another. Two generated representations may be mutually consistent while describing a state poorly supported by real data, or a model may successfully manipulate a requested phenotype while altering attributes intended to remain fixed.

We study this principle through text-conditioned generation of microscopy images and paired instance masks. Using LIVECell \cite{LIVECell}, we construct prompts describing cell identity, acquisition time, density, confluency, morphology, and nearest-neighbor distance, and introduce T2Joint, a system for joint microscopy image and instance-mask generation. This exposes the generated biological state through three representations: the requested description, microscopy appearance, and instance organization. We evaluate Prompt--Image, Prompt--Mask, and Image--Mask relationships together with the resulting phenotype states in the real population (Fig.~\ref{fig:fig1}D). We further stress the system with increasingly unsupported phenotype combinations to examine what happens when explicit control conflicts with relationships represented in real data.

Our contributions are threefold. First, we formulate \textit{biological congruence} as a relational evaluation principle emphasizing coherence across representations and empirical context. Second, we develop a text-conditioned joint microscopy image--instance-mask system with biological prompting and structured instance representation, enabling biological information to be evaluated across complementary representations. Third, we show that these evaluation perspectives reveal feature-dependent patterns of convergence and dissociation, and that successful target control does not necessarily preserve the complete biological state.

\section{Related Work}

\paragraph{Generative microscopy and biological evaluation.}
Generative models have been applied to microscopy synthesis and to modeling cellular phenotypes and responses to experimental perturbations \cite{MorphoDiff,CellFlux,GAN_review_plos}. Their evaluation has accordingly expanded beyond generic measures of visual fidelity and distributional similarity to morphological descriptors, phenotype distributions, perturbation recovery, and downstream biological tasks \cite{IMPA,MORPHGEN,CellFlux}. These approaches establish whether particular biological properties are preserved, but generally evaluate them individually.

\paragraph{Conditional and cross-representation consistency.}
Conditional generation additionally requires generated samples to express the information specified by their conditioning variables, motivating dedicated evaluation of prompt adherence and compositional control \cite{TIFA,T2ICompBench}. Biological information can also be distributed across complementary representations. Recent approaches align or jointly model cellular morphology, molecular measurements, spatial context, microscopy images, and segmentation masks \cite{Spatia,sccross,medsegfactory,MorphoDiff,icaart25}, highlighting the importance of relationships between representations of biological state. We connect these perspectives by treating conditioning, microscopy appearance, and instance structure as complementary observations of the same generated state.

\paragraph{Biological congruence.}
Existing strategies therefore evaluate complementary properties such as generative quality, conditional adherence, phenotype preservation, and cross-representation consistency. We formulate \textit{biological congruence} to examine these properties jointly with their empirical biological context. Rather than aggregating them into a single score, we study where these perspectives agree or dissociate, including when explicit phenotype control conflicts with relationships supported by real data.

\section{Biologically Congruent Microscopy Generation}
\label{sec:Methods}

\begin{figure*}[t]
  \centering
  \includegraphics[width=\linewidth]{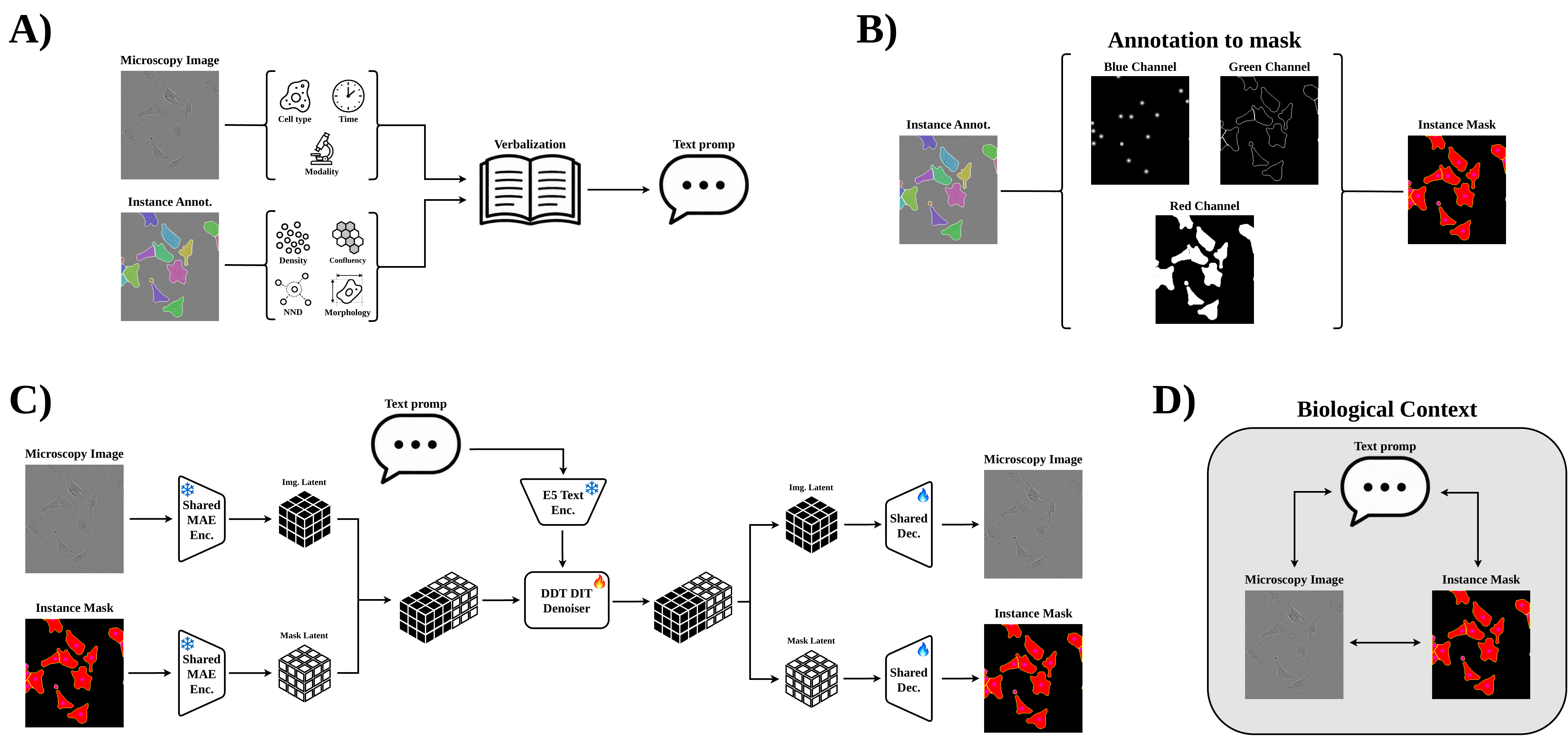}
  \caption{\textbf{Overview of biologically congruent microscopy generation and evaluation.}
  \textbf{A)} Biological metadata and annotation-derived phenotypes are converted into natural-language prompts through a predefined codebook.
  \textbf{B)} Instance annotations are represented by three channels encoding foreground occupancy, object boundaries, and instance centers.
  \textbf{C)} T2Joint jointly generates paired microscopy images and structured instance masks through a shared latent representation conditioned on the prompt.
  \textbf{D)} Biological congruence is evaluated through Prompt--Image, Prompt--Mask, and Image--Mask relationships, interpreted within the empirical biological context established by real data.
  }
  \label{fig:fig1}
\end{figure*}

\subsection{How do we represent and generate biological state?}
\label{sec:representation_generation}

Our framework represents biological information through text, microscopy appearance, and instance-level structure (Fig.~\ref{fig:fig1}A--B). We construct biological prompts from image metadata and annotation-derived phenotypes, transform instance annotations into a generative three-channel representation, and adapt a text-conditioned representation autoencoder (RAE) to jointly generate microscopy images and masks (Fig.~\ref{fig:fig1}C).

\paragraph{Dataset and biological representations.}
We use LIVECell~\cite{LIVECell}, a phase-contrast microscopy dataset comprising eight cell types (A172, BT-474, BV-2, Huh7, MCF7, SH-SY5Y, SkBr3, and SK-OV-3), more than 5,000 images acquired at 4-h intervals, and over 1.6 million annotated cells. Images are divided into six $260\times236$ tiles, padded to $260\times256$, and center-cropped to $256\times256$ for training.

\paragraph{Instance-mask representation.}
We transform each instance annotation into a three-channel representation encoding foreground occupancy, object boundaries, and a per-instance Gaussian-like field centered at the maximum-inscribed-disk center (Fig.~\ref{fig:fig1}B). These channels encode object extent, separation between adjacent instances, and an interior instance cue, respectively. The representation can be generated as a three-channel image and converted back into individual instances.

\paragraph{Biological conditioning.}
Each sample is described by seven attributes (Fig.~\ref{fig:fig1}A): cell type, acquisition time, microscopy modality, and four annotation-derived phenotypes. These comprise \textit{density} (cell count), \textit{confluency} (fraction of occupied image area), \textit{morphology} (cell area-to-convex-area ratio), and \textit{nearest-neighbor distance (NND)}.

The four phenotypes are converted into ordinal descriptions using cell-type-specific quantile stratification. Each phenotype and cell-type training distribution is divided into seven approximately balanced strata, with three synonymous descriptions assigned to each stratum through a predefined codebook. One description per phenotype is sampled into one of three prompt templates. Acquisition time remains numerical, while cell type and modality are inserted directly. Detailed feature definitions, codebooks, and templates are provided in the supplementary material.

\paragraph{Text-conditioned joint generation.}
T2Joint extends the RAE framework~\cite{RAE} to text-conditioned joint generation of microscopy images and structured instance masks (Fig.~\ref{fig:fig1}C). Prompts are represented by a frozen E5 encoder~\cite{E5}, whose text--text contrastive pretraining is well aligned with our ordinal textual conditioning compared with the image--text objective of CLIP~\cite{CLIP}. Text representations condition every other DDT encoder block through cross-attention~\cite{DDT_DiT}.

Images and masks are independently encoded by the same pretrained frozen MAE encoder~\cite{MAE}, and their latent representations are concatenated for joint denoising. The resulting denoised latent is split and independently decoded into image and mask components. Following RAE adaptation, the decoder is fine-tuned on microscopy images while the encoder remains frozen. Structured masks are not used during decoder adaptation but remain reconstructible by the resulting decoder. Training details are provided in the supplementary material.

RAE was selected as the generative backbone after comparison with Kandinsky and MedSegFactory, achieving the best FID, precision, and recall among the evaluated alternatives (see supplementary material).

\subsection{How do we evaluate biological congruence?}
\label{sec:biological_congruence}

We use two synthetic datasets. First, each prompt from the LIVECell validation split is generated once, producing 3,404 paired synthetic samples with matching validation conditions. Second, we construct a \textit{traversal dataset} from five anchors per cell type distributed across acquisition time. From these 40 anchors, each of the four mask-derived phenotypes is independently traversed through its six alternative strata while all other attributes remain fixed, yielding 960 traversal conditions. The complete traversal is generated with three independent seeds.

We report FID, generative precision, and recall~\cite{FID,recall_precision} from InceptionV3~\cite{inceptionv3} representations as conventional measures of image fidelity and distribution coverage. Biological congruence is then evaluated through relationships between prompt, image, mask, and the empirical context established by real data (Fig.~\ref{fig:fig1}D).

\paragraph{Does the generated mask express the requested biological state?}
For Prompt--Mask alignment, we recompute density, confluency, NND, and morphology from each generated mask and map them to the same cell-type-specific strata used for prompt construction. We quantify \textit{absolute alignment} between requested and realized strata using quadratic weighted Cohen's $\kappa$ (QWK)~\cite{QWK}. Additional agreement and rank-based measures are reported in the supplementary material.

We additionally measure \textit{directional control} using the traversal dataset. For feature $f$ and traversal $i$, directional success compares the sign of the requested change from anchor $a$ with the corresponding change in the generated mask:
\begin{equation}
\begin{aligned}
D_{f,i}
&=\mathbb{I}\!\left[
\operatorname{sgn}(\Delta S_{f,i}^{\mathrm{gen}})
=
\operatorname{sgn}(\Delta S_{f,i}^{\mathrm{prompt}})
\right],\\
D_f
&=\frac{1}{N_f}\sum_i D_{f,i}.
\end{aligned}
\end{equation}

Higher $D_f$ indicates that generated changes more consistently follow the requested direction. It evaluates direction rather than magnitude; anchors serve only as baselines and are excluded from the success rate.

\paragraph{Does the generated image express the requested biological state?}
Prompt--Image agreement is evaluated for cell type and acquisition time, the two varying image-associated prompt attributes. We encode each image with the frozen microSAM image encoder~\cite{microSAM}, a microscopy-specific adaptation of SAM~\cite{SAM}, and mean-pool the valid spatial region into a 256-dimensional representation. Lightweight probes trained exclusively on real LIVECell training representations are applied unchanged to real and synthetic validation images. Real-validation performance provides a reference for information recoverable by the representation and probe.

Cell type is evaluated using multinomial logistic regression and classification accuracy; acquisition time using Ridge regression and MAE in hours. Duplicate images shared by the original LIVECell training and validation splits are excluded from probe training to prevent leakage. Further probe details are provided in the supplementary material.

\paragraph{Do image and mask describe the same generated state?}
For Image--Mask agreement, an independent LIVECell instance-segmentation model~\cite{LIVECell} infers cellular structure from each generated image, which is compared with its jointly generated mask. The same procedure is applied to genuine LIVECell image--annotation pairs to characterize evaluator disagreement on real microscopy.

At the structural level, we report Panoptic Quality (PQ)~\cite{PQ} between the independently inferred and reference masks. For synthetic samples the generated mask is the reference; for real samples the ground-truth annotation is the reference. Real-data PQ provides a practical evaluator reference rather than an upper bound. Additional segmentation diagnostics are reported in the supplementary material.

At the phenotype level, density, confluency, NND, and morphology are independently extracted from the reference and inferred masks. Each feature is mapped to empirical percentiles using its cell-type-specific real-training distribution, and sample disagreement is
\begin{equation}
e_{i,f}=\left|p^{\mathrm{inf}}_{i,f}-p^{\mathrm{ref}}_{i,f}\right|.
\end{equation}
To account for evaluator error, synthetic disagreement is calibrated against genuine LIVECell pairs grouped by cell type and five cell-type-specific temporal bins:
\begin{equation}
E^{\mathrm{excess}}_{i,f}
=
e^{\mathrm{syn}}_{i,f}
-
\mathbb{E}\!\left[e^{\mathrm{real}}_f\mid c\right].
\end{equation}
Values near zero indicate disagreement comparable to the evaluator on matched genuine pairs; positive values indicate additional disagreement. Negative values are conservatively interpreted as no detectable excess disagreement. This calibration does not recover unknown synthetic ground truth.

\paragraph{Does the generated state preserve empirical biological context?}
We compare generated and real phenotypes at marginal and conditional levels. For \textit{marginal phenotype fidelity}, density, confluency, NND, and morphology are standardized using cell-type-specific real-training medians and IQRs. Their generated and real validation distributions are compared using one-dimensional Wasserstein distance ($W_1$), where lower values indicate closer marginal agreement.

For \textit{conditional phenotype structure}, each ordered phenotype pair $X\rightarrow Y$ is stratified by the real-derived ordinal levels of $X$. We compare the median standardized response $Y$ between generated and real samples within each valid stratum:
\begin{equation}
E_{X\rightarrow Y}
=
\frac{1}{|\mathcal{S}_X|}
\sum_{s\in\mathcal{S}_X}
\left|
\tilde{Y}^{\,\mathrm{gen}}_s-\tilde{Y}^{\,\mathrm{real}}_s
\right|.
\end{equation}
Lower values indicate closer preservation of the real conditional response. The measure is directional because $X$ defines the stratification. For the feature-level summary in Fig.~\ref{fig:fig2}, we average the three relationships in which $Y$ is the response:
\begin{equation}
E_{\mathrm{cond}}(Y)=\frac{1}{3}\sum_{X\neq Y}E_{X\rightarrow Y}.
\end{equation}

\paragraph{What happens when control conflicts with empirical biology?}
We use the traversal dataset to progressively move requested phenotype configurations away from states represented in real data. For each requested phenotype vector $\mathbf{v}$, we restrict the real training reference set to matching cell type and acquisition time and define
\begin{equation}
\begin{aligned}
d(\mathbf{v},\mathbf{x})
&=\frac{1}{4}\sum_{f=1}^{4}|v_f-x_f|,\\
D_5(\mathbf{v})
&=\frac{1}{5}\sum_{\mathbf{x}\in5\mathrm{NN}(\mathbf{v})}
d(\mathbf{v},\mathbf{x}).
\end{aligned}
\end{equation}

The same $D_5$ is computed for real training samples using leave-one-out neighbors. Traversals are assigned percentiles of this real-data distance distribution and grouped into $<80$, $80$--$90$, $90$--$95$, $95$--$99$, and $>99$ ranges, with higher percentiles indicating weaker empirical support.

We evaluate target control using directional success and quantify changes in the three attributes intended to remain fixed through
\begin{equation}
E_{\mathrm{off},i}
=
\frac{1}{3}
\sum_{j\neq f}
\left|S^{\mathrm{gen}}_{j,i}-S^{\mathrm{prompt}}_{j,i}\right|.
\end{equation}
Finally, we measure whether the realized state moves toward or away from the empirical population:
\begin{equation}
\Delta D_i =
D_5(\mathbf{v}^{\mathrm{gen}}_i)
-
D_5(\mathbf{v}^{\mathrm{prompt}}_i).
\end{equation}
Negative $\Delta D$ indicates movement toward configurations represented in the matched real population, while positive values indicate movement away from them. This measures relative empirical support and does not establish biological correctness.
\section{Results}
\label{sec:Results}

\begin{figure*}[t]
  \centering
   \includegraphics[width=.85\linewidth]{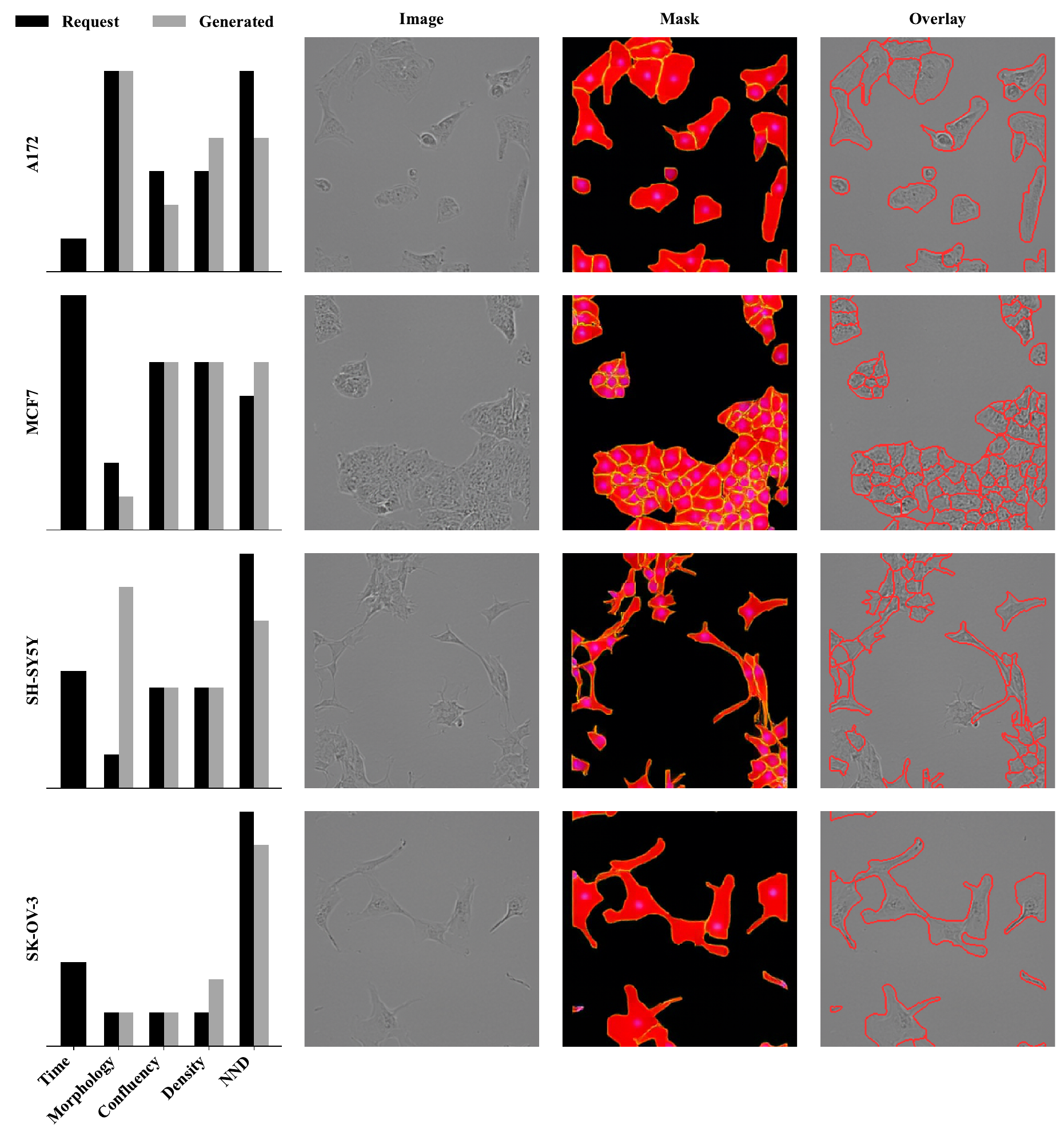}

    \caption{\textbf{Qualitative examples of text-conditioned joint microscopy image and instance-mask generation.}
    Each row shows, from left to right, requested and realized phenotype strata for morphology, confluency, density, and nearest-neighbor distance (NND); the generated microscopy image; the jointly generated structured instance mask; and their overlay.
    }
   \label{fig:fig4}
\end{figure*}

Figure~\ref{fig:fig4} shows representative outputs of T2Joint across cell types and biological conditions. The model jointly produces microscopy images and structured instance masks from the requested phenotype descriptions, with the overlay illustrating their spatial correspondence. The requested and realized phenotype strata further provide a qualitative view of how the generated state reflects the conditioning signal. We next evaluate these relationships quantitatively, beginning with conventional generative performance before examining biological congruence across representations and empirical context.

\subsection{Do conventional metrics tell the whole story?}
\label{sec:results_conventional}

We first examine whether conventional generative performance reflects the biological information preserved by generated images (Table~\ref{tab:FID_classifier}). The RAE backbone and text-conditioned image generator (T2I) achieve comparable conventional performance, with FID scores of $16.509$ and $16.131$, respectively, and identical recall ($0.349$). Extending generation to paired images and instance masks (T2Joint) incurs a measurable cost: FID increases to $20.853$, while precision and recall decrease from $0.801/0.349$ to $0.786/0.292$.

\begin{table}[t]
\centering
\scriptsize
\caption{Conventional generative performance and biological identity fidelity across model variants.}
\label{tab:FID_classifier}
\begin{tabular}{c|ccccc}
\hline
\textbf{Model} & \textbf{FID ↓}  & \textbf{Precision ↑} & \textbf{Recall ↑} & \textbf{\begin{tabular}[c]{@{}l@{}}Cell type\\ acc. ↑\end{tabular}} & \textbf{\begin{tabular}[c]{@{}l@{}}Time \\ MAE ↓\end{tabular}} \\ \hline
RAE            & 16.509          & 0.799                & 0.349             & 0.844                     & —                              \\
T2I            & \textbf{16.131} & \textbf{0.801}       & \textbf{0.349}    & \textbf{0.851}            & \textbf{14.925}                \\
T2Joint        & 20.853          & 0.786                & 0.292             & 0.844                     & 15.214                         \\ \hline
\end{tabular}
\end{table}

This deterioration is not mirrored by biological identity. Cell-type accuracy remains stable across RAE, T2I, and T2Joint ($0.844$, $0.851$, and $0.844$, respectively). T2I and T2Joint also recover acquisition time with similar MAEs of $14.925$ and $15.214$ h ($3.73$ and $3.80$ nominal 4-h acquisition intervals). Unlike the phenotype attributes represented through cell-type-specific ordinal strata, acquisition time is conditioned at its original numerical resolution across cell-type trajectories of different duration. For reference, the same frozen probe achieves $9.678$ h MAE on real validation images. Temporal recovery is therefore less precise than on genuine microscopy, but joint image--mask generation introduces little additional degradation relative to image-only generation.

These results show the complementary roles of conventional and biological evaluation. FID, precision, and recall expose a distributional cost associated with joint generation, but this cost is not proportionally reflected in the recoverability of cell type or acquisition time. Conversely, preservation of these global attributes cannot establish whether the richer biological state remains coherent across conditioning, image structure, instance annotations, and the empirical population.

\subsection{Is biological information preserved consistently across perspectives?}
\label{sec:results_congruence}

\begin{figure}[t]
  \centering
  \includegraphics[width=\linewidth]{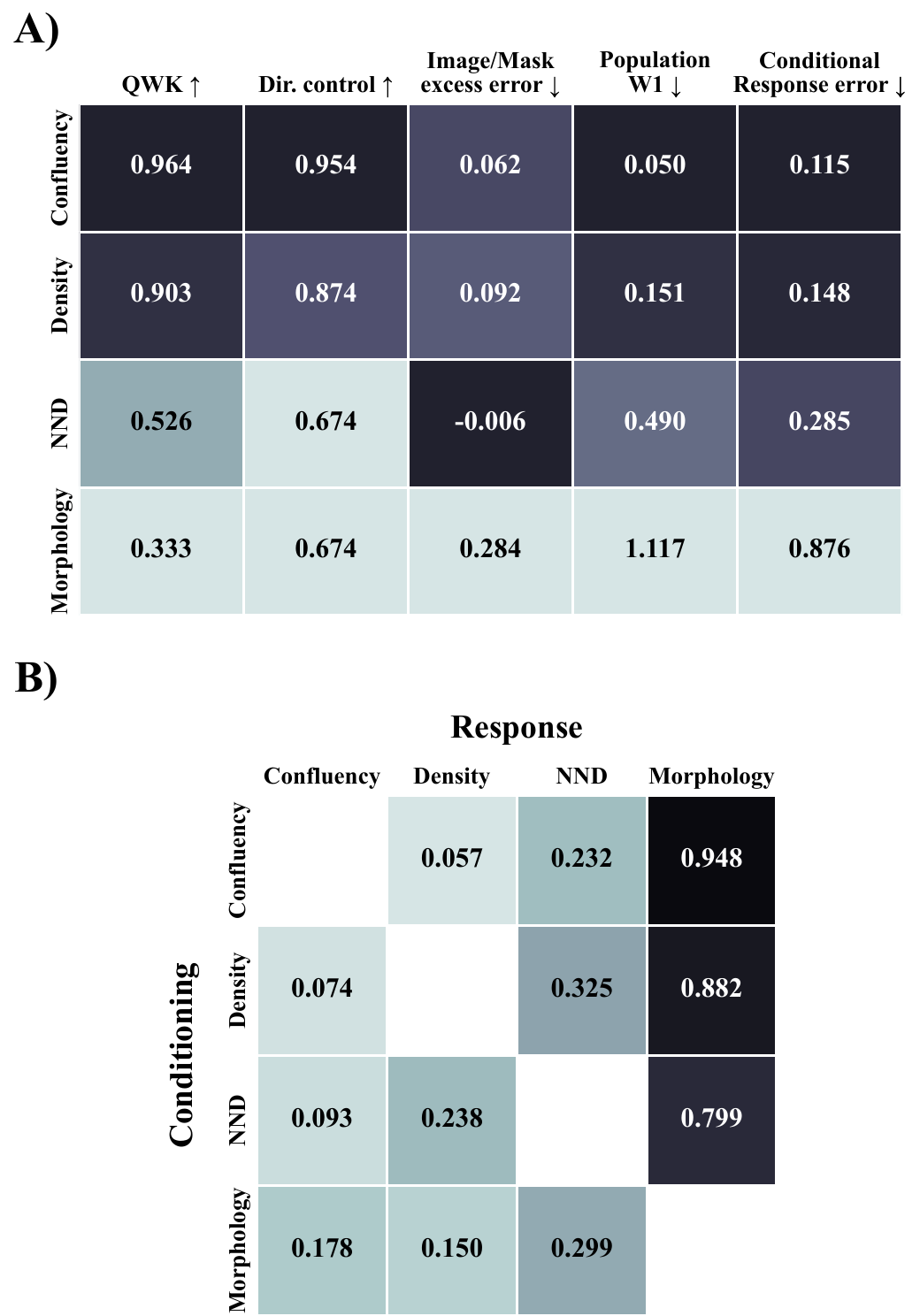}

  \caption{\textbf{Cross-perspective evaluation of biological phenotype preservation.}
  \textbf{A)} Morphology, confluency, density, and nearest-neighbor distance (NND) evaluated through prompt--mask alignment (QWK), directional control, calibrated image--mask phenotype disagreement, population Wasserstein distance, and conditional-response error. Shading is normalized independently within each perspective for visualization; values remain on their native scales and are not aggregated.
  \textbf{B)} Directed conditional-response error between phenotype pairs. Rows denote the conditioning phenotype and columns the response phenotype; values quantify generated--real differences in conditional median responses across real-derived strata.
  }

  \label{fig:fig2}
\end{figure}

Biological congruence reveals feature-dependent behavior that is not apparent from image-level metrics alone (Fig.~\ref{fig:fig2}A). Confluency and density show the most consistent preservation across perspectives. Their requested strata are accurately recovered from generated masks (QWK $=0.964/0.903$ for confluency/density), and controlled traversals predominantly move both phenotypes in the requested direction ($D_f=0.954/0.874$). Image-derived segmentations introduce only $0.062/0.092$ excess percentile error relative to the generated masks, while the generated populations remain close to the real marginal distributions ($W_1=0.050/0.151$) and reproduce their conditional phenotype behavior with low errors ($0.115/0.148$). Evidence from control, cross-representation agreement, and population context therefore converges for these attributes.

NND instead exposes a cross-perspective dissociation. Its prompt--mask agreement and directional control are lower (QWK $=0.526$, $D_f=0.674$), and its generated marginal distribution deviates more strongly from real data ($W_1=0.490$). Nevertheless, NND shows no detectable excess Image--Mask phenotype disagreement ($E_{\mathrm{excess}}=-0.006$). The paired image and mask can therefore express consistent NND information even when that information is less faithfully controlled from the prompt and reproduced at the population level. Agreement within one representation pair is thus insufficient to establish biological congruence of the complete generated state.

Morphology exhibits a different regime, with recurrent weakness across nearly every perspective. It has the lowest prompt--mask agreement (QWK $=0.333$), shares the lowest directional control with NND ($D_f=0.674$), and exhibits the largest Image--Mask excess error ($0.284$), marginal distribution discrepancy ($W_1=1.117$), and aggregate conditional-response error ($0.876$). The directed analysis further localizes this behavior (Fig.~\ref{fig:fig2}B): morphology is poorly reproduced as a response to density ($0.882$), confluency ($0.948$), and NND ($0.799$), while relationships in which morphology defines the conditioning strata are reproduced more closely. This asymmetry does not establish a causal or informational hierarchy, but shows that the generated population particularly fails to reproduce how morphology varies within its surrounding phenotype context. Because morphology is represented by a single area-to-convex-area descriptor, this recurrent failure may reflect limitations of the phenotype definition as well as the generator.

Together, these results expose three patterns: consistent preservation across representations and population context for density and confluency, cross-perspective dissociation for NND, and recurrent disagreement for morphology. These distinctions cannot be recovered from any single metric or representation relationship. At the structural level, generated image--mask pairs achieve PQ $=0.459$, compared with $0.577$ for genuine LIVECell image--annotation pairs under the same independent evaluator, indicating substantial but imperfect instance-level correspondence.

\subsection{What happens when control conflicts with empirical biology?}
\label{sec:results_conflict}

\begin{figure}[t]
  \centering
  \includegraphics[width=0.8\linewidth]{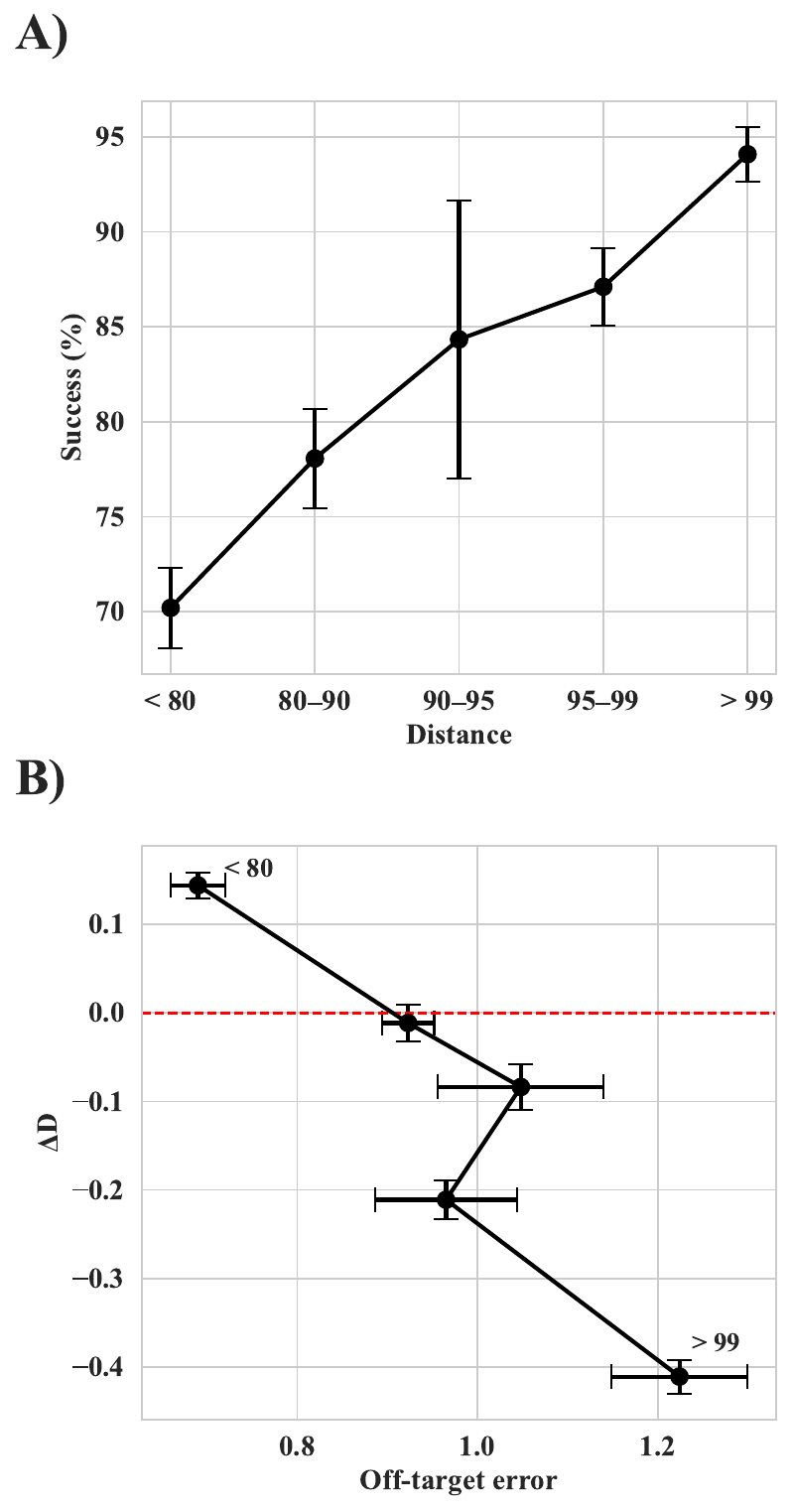}

  \caption{\textbf{Target control and surrounding-state preservation under increasing biological conflict.}
  \textbf{A)} Target directional success across empirical-support distance percentiles of the requested phenotype state.
  \textbf{B)} Off-target error and change in empirical-support distance ($\Delta D=D_{\mathrm{generated}}-D_{\mathrm{requested}}$) across the same bins. Negative $\Delta D$ indicates movement toward the matched real-data population. Points show means across three generation seeds; error bars denote $\pm1$ SD.
  }

  \label{fig:fig3}
\end{figure}

The traversal experiment reveals an unexpected relationship between explicit phenotype control and empirical biological support (Fig.~\ref{fig:fig3}). As requested phenotype combinations become increasingly distant from states observed in matched real data, directional success of the manipulated attribute increases monotonically from $0.702$ below the 80th distance percentile to $0.941$ beyond the 99th percentile. Increasingly unsupported requests therefore do not make the target phenotype less responsive to conditioning; instead, it follows the requested direction more consistently.

This increased target control does not imply preservation of the remaining biological state. Mean off-target error increases overall from $0.690$ below the 80th percentile to $1.224$ beyond the 99th percentile, indicating increasing disagreement among phenotypes requested to remain fixed. Target controllability and whole-state fidelity therefore become increasingly dissociated as the requested configuration conflicts with the empirical population.

This off-target reorganization does not move the complete generated state arbitrarily with respect to the real-data distribution. The relative empirical-support distance, $\Delta D$, changes from $+0.144$ below the 80th percentile to $-0.411$ beyond the 99th percentile (Fig.~\ref{fig:fig3}B). At extreme conflict, the phenotype state realized by the generator is therefore closer to configurations represented in matched real data than the state specified by the prompt, despite high directional success of the manipulated phenotype. Supplementary analyses further show that approximately $75\%$ of extreme traversals exhibit off-target changes aligned with phenotype directions observed in real data, while adjusted analyses indicate that the increase in target directional success is not explained solely by larger requested interventions.

These results expose a tension that target adherence alone cannot capture. Under increasing biological conflict, the generator can follow the requested direction of the manipulated phenotype while reorganizing untargeted attributes toward states with greater empirical support. Evaluating the complete generated state is therefore necessary to determine whether successful conditioning preserves, disrupts, or reorganizes its surrounding biological context.
\section{Discussion and Limitations}
\label{sec:Discussion}

Our experiments are limited to LIVECell, which was selected because its scale, multiple cell types, temporal sampling, and instance annotations enable conditioning, image appearance, instance structure, and population context to be evaluated within the same biological system. This allows a controlled multi-perspective analysis, but whether the observed patterns generalize to other microscopy datasets, modalities, or biological systems remains to be established.

Biological congruence depends on which biological properties are represented and how their relationships are evaluated. Morphology illustrates this dependence: the recurrent disagreement observed across perspectives concerns the area-to-convex-area descriptor used in our experiments and does not imply that morphological information in general cannot be preserved. Likewise, the Prompt--Image, Prompt--Mask, and Image--Mask relationships used here reflect our experimental setting and should not be interpreted as a universal evaluation protocol.

Our evaluation also relies on domain-specific proxies where synthetic ground truth is unavailable. microSAM representations and the LIVECell segmentation model provide independent references for information expressed in generated images, but remain imperfect estimators. We partially account for this limitation by training image probes only on real data and by calibrating image--mask phenotype disagreement against evaluator errors observed on genuine microscopy.

Finally, empirical support should not be interpreted as biological validity. The conflict experiment measures proximity to phenotype configurations observed in matched real data, not whether a generated state is biologically correct or mechanistically feasible. The observed reorganization of untargeted phenotypes therefore motivates further investigation of learned biological dependencies rather than evidence that the generator understands or corrects biological state. More generally, biological congruence is intended as a relational evaluation principle whose representations, biological properties, and empirical references should be chosen according to the scientific information a generative model is expected to preserve.

\section{Conclusion}
\label{sec:Conclusion}

We introduced \textit{biological congruence} as an evaluation principle for examining whether different representations of a generated sample provide coherent descriptions of its biological state and remain consistent with the empirical context they aim to represent. Using joint text-conditioned microscopy image and instance-mask generation as an experimental system, we showed that conventional generative quality, prompt adherence, cross-representation agreement, and population fidelity capture complementary rather than interchangeable properties. Across phenotypes, density and confluency show consistent preservation, NND reveals that image--mask agreement can coexist with weaker conditioning and population fidelity, and morphology exhibits recurrent disagreement across perspectives. Under increasing conflict with empirical support, target controllability can also increase while untargeted phenotypes reorganize toward states better represented in real data. These findings motivate evaluating biological generative models beyond aggregate quality or controllability metrics by considering the relationships among the biological information they are expected to preserve.

\section*{Acknowledgment}
This study was partially funded by Sartorius AI Lab (SAIL), a collaboration between Sartorius (\url{https://www.sartorius.com}) and the German Research Center for Artificial Intelligence (DFKI) (\url{https://www.dfki.de}). The funders had no role in study design, data collection and analysis, decision to publish, or preparation of the manuscript.

This work was also supported by Japan Science and Technology Agency (JST) as part of Adopting Sustainable Partnerships for Innovative Research Ecosystem (ASPIRE), Grant Number JPMJAP2403, and Japan Society for the Promotion of Science (JSPS) KAKENHI Grant Numbers JP24KJ1805.

{
    \small
    \bibliographystyle{ieeenat_fullname}
    \bibliography{main}

\begin{thebibliography}{27}
\providecommand{\natexlab}[1]{#1}
\providecommand{\url}[1]{\texttt{#1}}
\expandafter\ifx\csname urlstyle\endcsname\relax
  \providecommand{\doi}[1]{doi: #1}\else
  \providecommand{\doi}{doi: \begingroup \urlstyle{rm}\Url}\fi

\bibitem[Archit et~al.(2025)Archit, Freckmann, Nair, Khalid, Hilt, Rajashekar, Freitag, Teuber, Spitzner, Tapia~Contreras, Buckley, {von Haaren}, Gupta, Grade, Wirth, Schneider, Dengel, Ahmed, and Pape]{microSAM}
Anwai Archit, Luca Freckmann, Sushmita Nair, Nabeel Khalid, Paul Hilt, Vikas Rajashekar, Marei Freitag, Carolin Teuber, Melanie Spitzner, Constanza Tapia~Contreras, Genevieve Buckley, Sebastian {von Haaren}, Sagnik Gupta, Marian Grade, Matthias Wirth, G{\"u}nter Schneider, Andreas Dengel, Sheraz Ahmed, and Constantin Pape.
\newblock Segment {{Anything}} for {{Microscopy}}.
\newblock \emph{Nature Methods}, 22\penalty0 (3):\penalty0 579--591, 2025.

\bibitem[Cohen(1968)]{QWK}
Jacob Cohen.
\newblock Weighted kappa: {{Nominal}} scale agreement provision for scaled disagreement or partial credit.
\newblock \emph{Psychological Bulletin}, 70\penalty0 (4):\penalty0 213--220, 1968.

\bibitem[Cropsal and Mercado(2025)]{bio_eval}
T{\'e}lio Cropsal and Roc{\'{\i}}o Mercado.
\newblock Compressing biology: {{Evaluating}} the stable diffusion {{VAE}} for phenotypic drug discovery.
\newblock In \emph{{{NeurIPS}} 2025 Workshop for Imageomics: {{Discovering}} Biological Knowledge from Images Using {{AI}}}, 2025.

\bibitem[Demirel et~al.(2025)Demirel, Fumero, Karaletsos, and Locatello]{MORPHGEN}
Berker Demirel, Marco Fumero, Theofanis Karaletsos, and Francesco Locatello.
\newblock {{MorphGen}}: {{Controllable}} and morphologically plausible generative cell-imaging.
\newblock \emph{arXiv preprint arXiv:2510.01298}, 2025.

\bibitem[Edlund et~al.(2021)Edlund, Jackson, Khalid, Bevan, Dale, Dengel, Ahmed, Trygg, and Sj{\"o}gren]{LIVECell}
Christoffer Edlund, Timothy~R. Jackson, Nabeel Khalid, Nicola Bevan, Timothy Dale, Andreas Dengel, Sheraz Ahmed, Johan Trygg, and Rickard Sj{\"o}gren.
\newblock {{LIVECell}}---{{A}} large-scale dataset for label-free live cell segmentation.
\newblock \emph{Nature Methods}, 18\penalty0 (9):\penalty0 1038--1045, 2021.

\bibitem[Farahani et~al.(2026)Farahani, Di~Ieva, Coiera, and Liu]{trust_gap}
Somayeh Farahani, Antonio Di~Ieva, Enrico Coiera, and Sidong Liu.
\newblock The trust gap in generative medical imaging: {{Evidence}}, risks, and a roadmap toward responsible adoption.
\newblock \emph{European Journal of Radiology Artificial Intelligence}, 6:\penalty0 100092, 2026.

\bibitem[He et~al.(2022)He, Chen, Xie, Li, Dollar, and Girshick]{MAE}
Kaiming He, Xinlei Chen, Saining Xie, Yanghao Li, Piotr Dollar, and Ross Girshick.
\newblock Masked autoencoders are scalable vision learners.
\newblock In \emph{2022 {{IEEE}}/{{CVF}} Conference on Computer Vision and Pattern Recognition ({{CVPR}})}, pages 15979--15988, Los Alamitos, CA, USA, 2022. IEEE Computer Society.

\bibitem[Heusel et~al.(2017)Heusel, Ramsauer, Unterthiner, Nessler, and Hochreiter]{FID}
Martin Heusel, Hubert Ramsauer, Thomas Unterthiner, Bernhard Nessler, and Sepp Hochreiter.
\newblock {{GANs Trained}} by a {{Two Time-Scale Update Rule Converge}} to a {{Local Nash Equilibrium}}.
\newblock In \emph{Advances in {{Neural Information Processing Systems}}}. Curran Associates, Inc., 2017.

\bibitem[Hu et~al.(2023)Hu, Liu, Kasai, Wang, Ostendorf, Krishna, and Smith]{TIFA}
Yushi Hu, Benlin Liu, Jungo Kasai, Yizhong Wang, Mari Ostendorf, Ranjay Krishna, and Noah~A. Smith.
\newblock {{TIFA}}: {{Accurate}} and interpretable text-to-image faithfulness evaluation with question answering.
\newblock In \emph{2023 {{IEEE}}/{{CVF}} International Conference on Computer Vision ({{ICCV}})}, pages 20349--20360, 2023.

\bibitem[Huang et~al.(2023)Huang, Sun, Xie, Li, and Liu]{T2ICompBench}
Kaiyi Huang, Kaiyue Sun, Enze Xie, Zhenguo Li, and Xihui Liu.
\newblock {{T2I-CompBench}}: A comprehensive benchmark for open-world compositional text-to-image generation.
\newblock In \emph{Advances in Neural Information Processing Systems}, pages 78723--78747. Curran Associates, Inc., 2023.

\bibitem[Jiang et~al.(2024)Jiang, Song, Zhang, Huang, Deng, Sun, Zhang, Wang, and Zhuang]{E5}
Ting Jiang, Minghui Song, Zihan Zhang, Haizhen Huang, Weiwei Deng, Feng Sun, Qi Zhang, Deqing Wang, and Fuzhen Zhuang.
\newblock E5-v: {{Universal}} embeddings with multimodal large language models.
\newblock \emph{arXiv preprint arXiv:2407.12580}, 2024.

\bibitem[Kirillov et~al.(2019)Kirillov, He, Girshick, Rother, and Dollar]{PQ}
Alexander Kirillov, Kaiming He, Ross Girshick, Carsten Rother, and Piotr Dollar.
\newblock Panoptic segmentation.
\newblock In \emph{Proceedings of the {{IEEE}}/{{CVF}} Conference on Computer Vision and Pattern Recognition ({{CVPR}})}, 2019.

\bibitem[Kirillov et~al.(2023)Kirillov, Mintun, Ravi, Mao, Rolland, Gustafson, Xiao, Whitehead, Berg, Lo, Dollar, and Girshick]{SAM}
Alexander Kirillov, Eric Mintun, Nikhila Ravi, Hanzi Mao, Chloe Rolland, Laura Gustafson, Tete Xiao, Spencer Whitehead, Alexander~C. Berg, Wan-Yen Lo, Piotr Dollar, and Ross Girshick.
\newblock Segment anything.
\newblock In \emph{Proceedings of the {{IEEE}}/{{CVF}} International Conference on Computer Vision ({{ICCV}})}, pages 4015--4026, 2023.

\bibitem[Kong et~al.(2025)Kong, Qiu, Boesen, Lin, Yun, Chen, Kellis, and Zitnik]{Spatia}
Zhenglun Kong, Mufan Qiu, John Boesen, Xiang Lin, Sukwon Yun, Tianlong Chen, Manolis Kellis, and Marinka Zitnik.
\newblock {{SPATIA}}: {{Multimodal}} generation and prediction of spatial cell phenotypes.
\newblock \emph{arXiv preprint arXiv:2507.04704}, 2025.

\bibitem[{Lesmes-Leon} et~al.(2025{\natexlab{a}}){Lesmes-Leon}, Miranda, Caroprese, Lovell, Dengel, and Ahmed]{icaart25}
Duway {Lesmes-Leon}, Miro Miranda, Maria Caroprese, Gillian Lovell, Andreas Dengel, and Sheraz Ahmed.
\newblock Synthesizing annotated cell microscopy images with generative adversarial networks.
\newblock In \emph{Proceedings of the 17th International Conference on Agents and Artificial Intelligence - Volume 3: {{ICAART}}}, pages 592--599. SciTePress / INSTICC, 2025{\natexlab{a}}.

\bibitem[{Lesmes-Leon} et~al.(2025{\natexlab{b}}){Lesmes-Leon}, Dengel, and Ahmed]{GAN_review_plos}
Duway~Nicolas {Lesmes-Leon}, Andreas Dengel, and Sheraz Ahmed.
\newblock Systematic review of generative adversarial networks ({{GANs}}) in cell microscopy: {{Trends}}, practices, and impact on image augmentation.
\newblock \emph{PLOS ONE}, 20\penalty0 (6):\penalty0 e0291217, 2025{\natexlab{b}}.

\bibitem[{Lesmes-Leon} et~al.(2026){Lesmes-Leon}, Schmeisser, Lovell, Dengel, and Ahmed]{custom_FID}
Duway~Nicolas {Lesmes-Leon}, Fabian Schmeisser, Gillian Lovell, Andreas Dengel, and Sheraz Ahmed.
\newblock Is realism enough? {{Evaluating}} generative models in cell microscopy.
\newblock In \emph{Proceedings of the {{IEEE}}/{{CVF}} Conference on Computer Vision and Pattern Recognition ({{CVPR}}) Workshops}, pages 6515--6524, 2026.

\bibitem[Mao et~al.(2025)Mao, Wang, Tang, Xu, Wang, Yang, Zhou, and Zhou]{medsegfactory}
Jiawei Mao, Yuhan Wang, Yucheng Tang, Daguang Xu, Kang Wang, Yang Yang, Zongwei Zhou, and Yuyin Zhou.
\newblock {{MedSegFactory}}: {{Text-guided}} generation of medical image-mask pairs.
\newblock In \emph{Proceedings of the {{IEEE}}/{{CVF}} International Conference on Computer Vision ({{ICCV}})}, pages 21525--21535, 2025.

\bibitem[Navidi et~al.(2025)Navidi, Ma, Miglietta, Liu, Carpenter, Cimini, {Haibe-Kains}, and WANG]{MorphoDiff}
Zeinab Navidi, Jun Ma, Esteban Miglietta, Le Liu, Anne Carpenter, Beth Cimini, Benjamin {Haibe-Kains}, and {\relax BO} WANG.
\newblock {{MorphoDiff}}: {{Cellular}} morphology painting with diffusion models.
\newblock In \emph{International Conference on Learning Representations}, pages 11613--11637, 2025.

\bibitem[Palma et~al.(2025)Palma, Theis, and Lotfollahi]{IMPA}
Alessandro Palma, Fabian~J. Theis, and Mohammad Lotfollahi.
\newblock Predicting cell morphological responses to perturbations using generative modeling.
\newblock \emph{Nature Communications}, 16\penalty0 (1):\penalty0 505, 2025.

\bibitem[Radford et~al.(2021)Radford, Kim, Hallacy, Ramesh, Goh, Agarwal, Sastry, Askell, Mishkin, Clark, et~al.]{CLIP}
Alec Radford, Jong~Wook Kim, Chris Hallacy, Aditya Ramesh, Gabriel Goh, Sandhini Agarwal, Girish Sastry, Amanda Askell, Pamela Mishkin, Jack Clark, et~al.
\newblock Learning transferable visual models from natural language supervision.
\newblock In \emph{International Conference on Machine Learning}, pages 8748--8763. PmLR, 2021.

\bibitem[Sajjadi et~al.(2018)Sajjadi, Bachem, Lucic, Bousquet, and Gelly]{recall_precision}
Mehdi S.~M. Sajjadi, Olivier Bachem, Mario Lucic, Olivier Bousquet, and Sylvain Gelly.
\newblock Assessing generative models via precision and recall.
\newblock In \emph{Advances in Neural Information Processing Systems}. Curran Associates, Inc., 2018.

\bibitem[Szegedy et~al.(2016)Szegedy, Vanhoucke, Ioffe, Shlens, and Wojna]{inceptionv3}
Christian Szegedy, Vincent Vanhoucke, Sergey Ioffe, Jon Shlens, and Zbigniew Wojna.
\newblock Rethinking the {{Inception Architecture}} for {{Computer Vision}}.
\newblock In \emph{2016 {{IEEE Conference}} on {{Computer Vision}} and {{Pattern Recognition}} ({{CVPR}})}, pages 2818--2826, 2016.

\bibitem[Wang et~al.(2026)Wang, Tian, Huang, and Wang]{DDT_DiT}
Shuai Wang, Zhi Tian, Weilin Huang, and Limin Wang.
\newblock {{DDT}}: {{Decoupled}} diffusion transformer.
\newblock In \emph{Proceedings of the {{IEEE}}/{{CVF}} Conference on Computer Vision and Pattern Recognition ({{CVPR}})}, pages 40633--40642, 2026.

\bibitem[Yang et~al.(2024)Yang, Mann, Wu, and Ding]{sccross}
Xiuhui Yang, Koren~K. Mann, Hao Wu, and Jun Ding.
\newblock {{scCross}}: A deep generative model for unifying single-cell multi-omics with seamless integration, cross-modal generation, and in silico exploration.
\newblock \emph{Genome Biology}, 25\penalty0 (1):\penalty0 198, 2024.

\bibitem[Zhang et~al.(2025)Zhang, Su, Wang, Li, Wefers, Nirschl, Burgess, Ding, Lozano, Lundberg, and {Yeung-Levy}]{CellFlux}
Yuhui Zhang, Yuchang Su, Chenyu Wang, Tianhong Li, Zoe Wefers, Jeffrey Nirschl, James Burgess, Daisy Ding, Alejandro Lozano, Emma Lundberg, and Serena {Yeung-Levy}.
\newblock {{CellFlux}}: {{Simulating}} cellular morphology changes via flow matching.
\newblock In \emph{International Conference on Machine Learning ({{ICML}})}, 2025.

\bibitem[Zheng et~al.(2026)Zheng, Ma, Tong, and Xie]{RAE}
Boyang Zheng, Nanye Ma, Shengbang Tong, and Saining Xie.
\newblock Diffusion transformers with representation autoencoders.
\newblock In \emph{International Conference on Learning Representations}, pages 35791--35820, 2026.

\end{thebibliography}
}

\clearpage
\appendix
\maketitlesupplementary

\renewcommand{\thesection}{\arabic{section}}
\setcounter{section}{2}

\section{Biologically Congruent Microscopy Generation}

\subsection*{Cellular Morphology and Spatial Organization Features}

Features are extracted from the instance masks. Let $N$ denote the number of cell instances, $A_i$ the segmented area of cell $i$, $H_i$ its convex hull area, and $A_{\mathrm{img}}$ the effective image area.

\paragraph{Morphology.}
Morphology is represented by the median cell solidity. For each cell,
\[
m_i=\frac{A_i}{H_i},
\]
and the image-level feature is
\[
\mathrm{Morphology}=\mathrm{median}_{i=1,\ldots,N}(m_i).
\]
Higher values correspond to more compact or convex shapes, while lower values indicate more irregular or concave shapes.

\paragraph{Confluency.}
Confluency is the fraction of the effective image area occupied by segmented cells:
\[
\mathrm{Confluency}
=
\frac{\left|\bigcup_{i=1}^{N}M_i\right|}{A_{\mathrm{img}}},
\]
where $M_i$ is the binary mask of cell $i$.

\paragraph{Density.}
Density is the number of detected cell instances:
\[
\mathrm{Density}=N.
\]

\paragraph{Nearest-neighbor distance (NND).}
Let $\mathbf{c}_i$ denote the centroid of cell $i$. Its nearest-neighbor distance is
\[
d_i=\min_{j\neq i}\left\|\mathbf{c}_i-\mathbf{c}_j\right\|_2,
\]
and the image-level NND is
\[
\mathrm{NND}=\mathrm{median}_{i=1,\ldots,N}(d_i).
\]
Smaller values indicate closer local cell spacing and larger values greater separation. NND is defined for images containing at least two cell instances.

\subsection*{Feature Stratum Codebook and Prompt Construction}

For each cell type, the training distribution of each phenotype is divided into seven quantile strata. A predefined codebook maps each stratum to three synonymous natural-language descriptions, resulting in $4\times7\times3=84$ phenotype descriptions. The complete codebook is provided in the accompanying \texttt{codebook.json} file.

During prompt construction, one description is sampled for each phenotype and inserted into one of three templates:

\begin{itemize}
  \item ``A \texttt{<modality>} microscopy image of \texttt{<cell type>} cells cultured for \texttt{<time>} hours. The image shows \texttt{<density>}, \texttt{<confluency>}, \texttt{<morphology>}, and \texttt{<NND>}.''
  
  \item ``\texttt{<modality>} microscopy showing \texttt{<cell type>} cells after \texttt{<time>} hours of culture. The tile contains \texttt{<density>}, with \texttt{<confluency>}, \texttt{<morphology>}, and \texttt{<NND>}.''
  
  \item ``A \texttt{<modality>} microscopy tile of \texttt{<cell type>} cells at \texttt{<time>} hours. The field is characterized by \texttt{<density>}, \texttt{<confluency>}, \texttt{<morphology>}, and \texttt{<NND>}.''
\end{itemize}

Example prompts include:

\begin{itemize}
  \item ``A phase-contrast microscopy tile of BT-474 cells at 4 hours. The field is characterized by a low number of cells, extensive visible background, extremely regular cell morphology, and extremely close neighboring cells.''
  
  \item ``phase-contrast microscopy showing SK-OV-3 cells after 24 hours of culture. The tile contains a sparse cell population, with considerable visible background, complex cell outlines, and large cell-cell spacing.''
  
  \item ``A phase-contrast microscopy image of SH-SY5Y cells cultured for 4 hours. The image shows relatively few cells, largely uncovered background, smooth cell outlines, and very close neighboring cells.''
\end{itemize}

\subsection*{Model and Training Details}

T2Joint uses a frozen RAE with a ViT-MAE-base encoder and a fine-tuned ViT-XL decoder. Microscopy images and structured instance masks are independently encoded into $768\times16\times16$ latent representations using the same encoder and separate latent normalization statistics. The representations are concatenated channel-wise into a $1536\times16\times16$ joint latent, jointly denoised, and split before independent decoding. Image and mask denoising losses are equally weighted.

The stage-2 denoiser uses DiTwDDTHead with encoder and decoder widths of $384$ and $2048$, 12 encoder blocks, 2 decoder blocks, and 6/16 attention heads. Text is encoded with a frozen E5-base-v2 encoder and projected to the denoiser representation. Token-level text cross-attention is applied every other encoder block, resulting in six cross-attention layers, with no additional text cross-attention in the decoder. Cross-attention residual gates are initialized to $0.1$. The class-conditioning pathway is disabled so that conditioning is provided exclusively through text.

\begin{table}[ht]
\centering
\small
\caption{Main T2Joint training configuration.}
\begin{tabular}{ll}
\toprule
\textbf{Parameter} & \textbf{Value} \\
\midrule
Image/mask latent size & $768\times16\times16$ each \\
Joint latent size & $1536\times16\times16$ \\
DDT hidden size & $[384,\,2048]$ \\
DDT depth & $[12,\,2]$ \\
Attention heads & $[6,\,16]$ \\
MLP ratio & $4.0$ \\
Text encoder & E5-base-v2 (frozen) \\
Text cross-attention & Every other encoder block (6) \\
Text dropout & $0.1$ \\
Global batch size & $128$ \\
Backbone learning rate & $2\times10^{-4}$ \\
Text-pathway learning rate & $10^{-3}$ \\
Gradient clipping & $1.0$ \\
EMA decay & $0.9995$ \\
Transport & Linear, velocity prediction \\
Time distribution & Logit-normal $(0,1)$ \\
Image/mask loss weighting & $1:1$ \\
Sampler & Heun, 50 steps \\
\bottomrule
\end{tabular}
\end{table}

T2Joint is trained from scratch using AdamW. Training is continued until the training loss and validation FID stabilize.

\subsection*{Preliminary Generative Backbone Selection}

Before developing T2Joint, we compared RAE with Kandinsky and MedSegFactory as candidate generative backbones. Table~\ref{tab:supp_backbone_selection} reports FID, precision, and recall for the three alternatives.

\begin{table}[ht]
\centering
\caption{Conventional generative performance of the models evaluated during preliminary backbone selection.}
\label{tab:supp_backbone_selection}
\begin{tabular}{lccc}
\toprule
\textbf{Model} & \textbf{FID $\downarrow$} & \textbf{Precision $\uparrow$} & \textbf{Recall $\uparrow$} \\
\midrule
Kandinsky     & 70.756          & 0.174          & 0.105 \\
MedSegFactory & 70.620          & 0.227          & \textbf{0.612} \\
RAE           & \textbf{16.509} & \textbf{0.799} & 0.349 \\
\bottomrule
\end{tabular}
\end{table}

\begin{figure*}[ht]
    \centering
    \includegraphics[width=\textwidth]{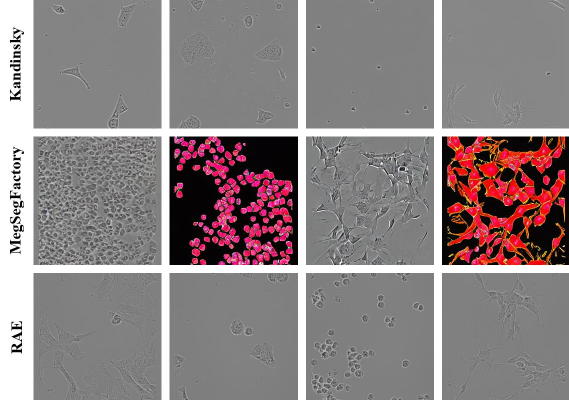}
    \caption{\textbf{Example generations from the models evaluated during preliminary backbone selection.}
    Four generated microscopy images are shown for Kandinsky and RAE. For MedSegFactory, two generated image--mask pairs are shown to additionally illustrate the quality of its jointly generated outputs. MedSegFactory examples exhibit visible image artifacts and incomplete image--mask correspondence, including cell instances visible in the generated images without corresponding instances in the generated masks. Examples were qualitatively selected to illustrate observed output characteristics.}
    \label{fig:supp_backbone_examples}
\end{figure*}

RAE substantially outperforms both alternatives in FID and precision. MedSegFactory achieves the highest recall ($0.612$), but this does not coincide with comparable generative fidelity: its FID and precision are $70.620$ and $0.227$, compared with $16.509$ and $0.799$ for RAE. Qualitative inspection further reveals visible artifacts in MedSegFactory images and incomplete correspondence between generated images and masks (Fig.~\ref{fig:supp_backbone_examples}). RAE was therefore selected based on its stronger overall generative fidelity rather than recall alone.

\section*{Prompt--Mask Biological Alignment}

\subsection*{Absolute Phenotype Alignment}

Complementary Prompt--Mask alignment measures are reported in Table~\ref{tab:supp_prompt_mask}. Phenotypes are recomputed directly from each generated mask and assigned to the same cell-type-specific strata used for prompt construction. In addition to quadratic weighted Cohen's $\kappa$ (QWK), reported in the main paper, we measure exact-stratum agreement, agreement within one stratum, and within-cell-type Spearman correlation between requested and realized strata.

\begin{table}[ht]
\small
\centering
\caption{Prompt--mask phenotype alignment on the synthetic validation set ($n=3404$).}
\label{tab:supp_prompt_mask}
\begin{tabular}{lrrrr}
\toprule
\textbf{Phenotype} & \textbf{QWK} & \textbf{Exact} & \textbf{$\pm1$ stratum} & \textbf{Spearman} \\
\midrule
Morphology  & 0.333 & 0.243 & 0.484 & 0.345 \\
Confluency  & 0.964 & 0.736 & 0.991 & 0.824 \\
Density     & 0.903 & 0.563 & 0.910 & 0.759 \\
NND         & 0.526 & 0.295 & 0.619 & 0.380 \\
\bottomrule
\end{tabular}
\end{table}

The complementary measures reproduce the hierarchy observed with QWK. Confluency and density show the strongest agreement, whereas NND and morphology are less accurately recovered from the requested strata.

\subsection*{Controlled Phenotype Traversals}

Directional control is evaluated using 40 real training anchors, with five anchors distributed across acquisition time for each of the eight cell types. Each phenotype is independently changed to its six alternative strata while all remaining prompt attributes are held fixed, producing $40\times4\times6=960$ traversal conditions. The complete traversal is generated with three independent seeds. For each traversal, the generated anchor from the same seed provides the reference state.

\begin{table}[ht]
\centering
\caption{Prompt--mask alignment under controlled phenotype traversals. Values are reported as mean $\pm$ SD across three independent generation seeds.}
\label{tab:supp_prompt_mask_traversal}
\begin{tabular}{lcc}
\toprule
\textbf{Phenotype} & \textbf{Directional success} & \textbf{Spearman} \\
\midrule
Morphology & $0.674 \pm 0.010$ & $0.384 \pm 0.026$ \\
Confluency & $0.954 \pm 0.011$ & $0.870 \pm 0.017$ \\
Density    & $0.874 \pm 0.027$ & $0.799 \pm 0.019$ \\
NND        & $0.674 \pm 0.054$ & $0.332 \pm 0.169$ \\
\bottomrule
\end{tabular}
\end{table}

Confluency and density remain substantially more controllable than morphology and NND when individual phenotype attributes are explicitly manipulated while the surrounding prompt is held fixed.

\section*{Prompt--Image Biological Congruence}

\subsection*{Image-Space Biological Probes}

Prompt--Image congruence is evaluated for cell type and acquisition time using representations from a frozen microSAM image encoder. The valid spatial region of each representation is mean-pooled to obtain a 256-dimensional feature vector. A multinomial logistic-regression classifier is used for cell type and Ridge regression for acquisition time. The feature scaler and both probes are fitted exclusively on real LIVECell training images and subsequently applied unchanged to real and synthetic validation images.

To avoid train--validation leakage in the probe evaluation, exact training images duplicated in the validation set are removed before probe fitting. This removes 178 images from the original 28,151-image training set, leaving 27,973 images for probe training. The validation set contains 3,404 images.

\subsection*{Prompt--Image Alignment}

Table~\ref{tab:supp_prompt_image} reports the complete probe results. Performance on genuine validation microscopy provides a reference for the biological information recoverable by each probe. Cell type is almost perfectly recovered from real images, with $0.996$ accuracy and macro-F1. Acquisition time is also strongly ordered ($\rho=0.894$), although its recovery is less precise, with an MAE of $9.678$ h.

\begin{table*}[ht]
\centering
\caption{Prompt--image biological information recovered from frozen microSAM representations. Probes are trained exclusively on real LIVECell training images and applied unchanged to real and synthetic validation images.}
\label{tab:supp_prompt_image}
\begin{tabular}{lrrrrrrr}
\toprule
& \multicolumn{3}{c}{\textbf{Cell type}} &
\multicolumn{4}{c}{\textbf{Acquisition time}} \\
\cmidrule(lr){2-4}\cmidrule(lr){5-8}
\textbf{Dataset}
& \textbf{Acc.}
& \textbf{Bal. Acc.}
& \textbf{Macro-F1}
& \textbf{MAE (h)}
& \textbf{RMSE (h)}
& \textbf{Spearman}
& $\mathbf{R^2}$ \\
\midrule
Real    & 0.996 & 0.995 & 0.996 &  9.678 & 12.703 & 0.894 & 0.779 \\
T2I     & 0.851 & 0.855 & 0.845 & 14.925 & 19.149 & 0.836 & 0.499 \\
T2Joint & 0.844 & 0.849 & 0.836 & 15.214 & 19.594 & 0.825 & 0.475 \\
\bottomrule
\end{tabular}
\end{table*}

T2I and T2Joint retain similar levels of biological identity information. Cell-type accuracy changes only from $0.851$ to $0.844$ after introducing joint image--mask generation, while time MAE changes from $14.925$ to $15.214$ h. The secondary metrics show the same pattern: balanced accuracy and macro-F1 differ by $0.006$ and $0.009$, respectively, while temporal Spearman correlation decreases from $0.836$ to $0.825$ and $R^2$ from $0.499$ to $0.475$.

\subsection*{Probe Diagnostics}

The reduction in T2Joint cell-type accuracy is structured across cell types rather than uniform. The largest confusion occurs for BT-474, which is frequently classified as Huh7 or MCF7, while several other cell types remain substantially more distinguishable. This contrasts with the near-perfect classification of genuine validation images and indicates that the aggregate decrease in synthetic cell-type accuracy is driven disproportionately by particular cellular identities.

Acquisition-time errors require additional calibration because temporal prediction is imperfect even on genuine microscopy. The real-data probe preserves temporal ordering strongly but increasingly underestimates later acquisition times. For example, mean predictions for real images acquired at 96, 108, and 120 h are $78.0$, $84.5$, and $87.6$ h, respectively. T2Joint exhibits the same general late-time compression, with corresponding mean predictions of $64.9$, $78.1$, and $90.1$ h. Temporal errors in synthetic images therefore occur on top of an existing limitation in recovering acquisition time from real microscopy representations rather than reflecting a failure unique to generated images.

\section*{Image--Mask Biological Congruence}

\subsection*{Structural Image--Mask Agreement}

Structural agreement is evaluated by applying an independent LIVECell instance-segmentation model to each microscopy image and comparing the inferred segmentation with its paired instance mask. The same evaluation is applied to genuine LIVECell image--annotation pairs to characterize the disagreement introduced by the evaluator itself.

\begin{table}[ht]
\centering
\footnotesize
\caption{Structural image--mask agreement for genuine LIVECell image--annotation pairs and T2Joint-generated image--mask pairs. FP and FN denote the mean numbers of false-positive and false-negative instances per image under the independent segmentation evaluator.}
\label{tab:supp_image_mask_structural}
\begin{tabular}{lrrrrrr}
\toprule
\textbf{Dataset} & \textbf{Dice $\uparrow$} & \textbf{PQ $\uparrow$} & \textbf{SQ $\uparrow$} & \textbf{RQ $\uparrow$} & \textbf{FP $\downarrow$} & \textbf{FN $\downarrow$} \\
\midrule
Real    & 0.840 & 0.577 & 0.791 & 0.720 & 10.47 & 18.47 \\
T2Joint & 0.776 & 0.459 & 0.747 & 0.598 & 11.77 & 23.87 \\
\bottomrule
\end{tabular}
\end{table}

T2Joint-generated pairs show lower structural agreement than genuine image--annotation pairs under the same evaluator. The difference is larger for recognition quality (RQ; $0.720$ vs.\ $0.598$) than for segmentation quality of matched instances (SQ; $0.791$ vs.\ $0.747$), indicating that instance correspondence contributes more strongly to the reduction in PQ than the segmentation quality of successfully matched objects. The evaluator itself exhibits non-zero disagreement on genuine microscopy, including more false-negative than false-positive instances. Relative to the real reference, T2Joint decreases Dice by $0.064$ (95\% bootstrap CI: $[-0.071,-0.057]$) and PQ by $0.118$ (95\% CI: $[-0.126,-0.109]$).

\subsection*{Phenotype-Level Image--Mask Agreement}

Phenotype-level disagreement is calibrated against the error observed when the same evaluator is applied to genuine microscopy. Density, confluency, morphology, and NND are independently computed from the paired reference mask and the image-derived segmentation. Continuous phenotype values are mapped to cell-type-specific empirical percentiles estimated from real training data. Absolute percentile disagreement is then calibrated against genuine image--annotation pairs matched by cell type and one of five cell-type-specific temporal bins.

Uncertainty is estimated using 10,000 bootstrap replicates. Real and synthetic observations are independently resampled and the condition-specific calibration is recomputed at each replicate. Table~\ref{tab:supp_image_mask_phenotype} reports the resulting excess disagreement and 95\% bootstrap confidence intervals.

\begin{table}[ht]
\centering
\caption{Calibrated phenotype-level image--mask disagreement for T2Joint. Excess error measures disagreement beyond that observed for genuine microscopy under the same evaluator. Values are reported with 95\% bootstrap confidence intervals.}
\label{tab:supp_image_mask_phenotype}
\begin{tabular}{lcc}
\toprule
\textbf{Phenotype} & \textbf{Excess error $\approx 0$} & \textbf{95\% CI} \\
\midrule
Morphology &  0.284 & $[0.273,\ 0.295]$ \\
Confluency &  0.062 & $[0.057,\ 0.067]$ \\
Density    &  0.092 & $[0.085,\ 0.098]$ \\
NND        & -0.006 & $[-0.014,\ 0.003]$ \\
\bottomrule
\end{tabular}
\end{table}

Confluency and density exhibit relatively small additional disagreement beyond the real-data evaluator baseline, whereas morphology shows substantially greater excess disagreement. For NND, the confidence interval includes zero, providing no evidence of additional image--mask disagreement beyond the level observed on genuine microscopy. Negative excess error should not be interpreted as synthetic pairs outperforming genuine annotations; it indicates that no excess disagreement is detected relative to the calibrated evaluator baseline.

\section{Biological-Conflict Stress Test}

\subsection*{Adjusted Target Directional Success}

Because increasingly unsupported requests may also involve larger phenotype changes, we tested whether the association between empirical-support distance and target directional success persists after accounting for intervention magnitude. For each generation seed, we fitted a logistic regression with target directional success as the binary outcome and standardized empirical-support distance and intervention magnitude as predictors, while additionally controlling for the traversed phenotype and cell type.

Empirical-support distance remained positively associated with directional success across all three generation seeds. A one-standard-deviation increase in distance was associated with $1.65$--$1.94$ times higher odds of directional success (seed-specific 95\% CIs: $[1.38,2.14]$, $[1.33,2.03]$, and $[1.52,2.49]$; all $p<4\times10^{-6}$). The increase in target directional success observed at greater biological conflict is therefore not explained solely by larger requested phenotype changes.

\subsection*{Direction of Off-Target Phenotype Reorganization}

We additionally examined whether off-target phenotype changes at extreme conflict follow directions observed in real data. For a traversal from requested stratum $s_0$ to $s_1$ of phenotype $f$, we estimate the expected response of the three untargeted phenotypes from their corresponding conditional medians in real training data. The expected off-target change is

\[
\Delta\mathbf{x}_{\mathrm{real}}
=
\mathbf{m}_{\mathrm{real}}(s_1)
-
\mathbf{m}_{\mathrm{real}}(s_0),
\]

where $\mathbf{m}_{\mathrm{real}}(s)$ contains the median standardized values of the three untargeted phenotypes among real samples in stratum $s$ of the traversed phenotype. The realized off-target change is computed relative to the generated anchor from the same seed. Directional alignment is quantified by the cosine similarity between the expected and realized off-target change vectors.

For requests beyond the 99th percentile of empirical-support distance, $74.9\%\pm3.0\%$ of traversals have positive cosine similarity across the three generation seeds. Mean cosine similarity is $0.348\pm0.022$, with a median of $0.507\pm0.023$. Thus, at extreme biological conflict, off-target phenotype changes preferentially follow directions consistent with phenotype relationships observed in real data. This directional analysis complements the negative $\Delta D$ observed in Fig. 4: the generated state not only moves toward configurations with greater empirical support, but its untargeted phenotype changes frequently align with directions present in the real population. These results characterize empirical alignment and should not be interpreted as evidence of biological correctness or model-level correction.

\section*{Qualitative results}

\begin{figure*}[ht]
    \centering
    \includegraphics[width=\textwidth]{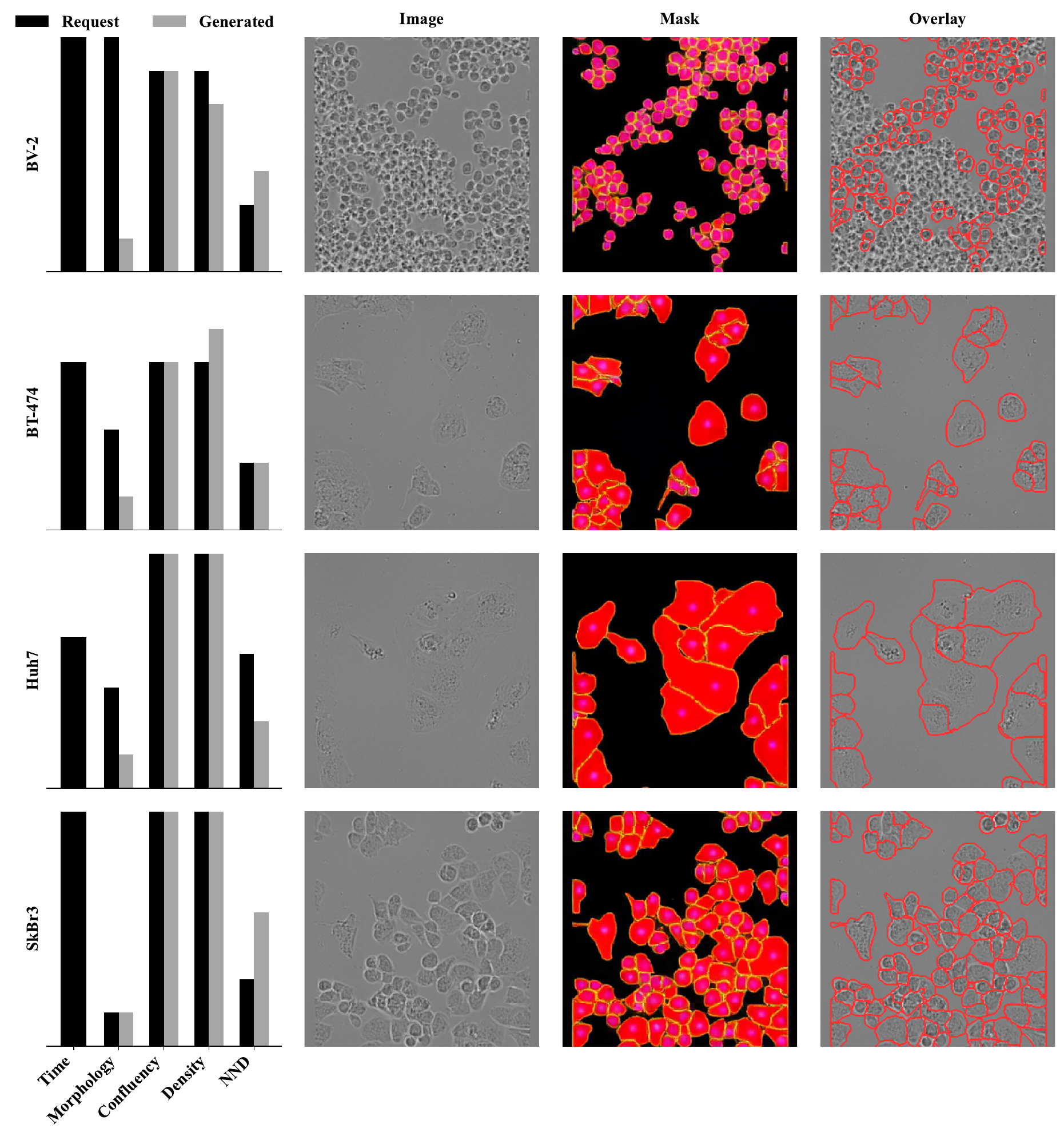}
    \caption{\textbf{Additional qualitative examples of text-conditioned joint microscopy image and instance-mask generation.}
    Each row shows, from left to right, requested and realized phenotype strata for morphology, confluency, density, and nearest-neighbor distance (NND); the generated microscopy image; the jointly generated structured instance mask; and their overlay. Examples span additional cell types, acquisition times, and phenotype configurations.}
    \label{fig:supp_qualitative}
\end{figure*}

\end{document}